\documentclass[conference]{IEEEtran}
\IEEEoverridecommandlockouts
\usepackage{cite}
\usepackage{amsmath,amssymb,amsfonts}
\usepackage{algorithmic}
\usepackage{graphicx}
\usepackage{textcomp}
\def\BibTeX{{\rm B\kern-.05em{\sc i\kern-.025em b}\kern-.08em
    T\kern-.1667em\lower.7ex\hbox{E}\kern-.125emX}}
    
\usepackage{hyperref}

\usepackage{caption}
\usepackage{subcaption}
\begin{document}

\title{Evaluation of Object Trackers in Distorted Surveillance Videos\\
\thanks{Colciencias - Pontificia Universidad Javeriana Colombia}
}

\author{\IEEEauthorblockN{Roger Gomez Nieto}
\IEEEauthorblockA{\textit{Department of Electronics and Computing} \\
\textit{Pontificia Universidad Javeriana}\\
Cali, Colombia\\
rogergo@javerianacali.edu.co}
\and
\IEEEauthorblockN{H.D. Benitez-Restrepo}
\IEEEauthorblockA{\textit{Department of Electronics and Computing} \\
\textit{Pontificia Universidad Javeriana}\\
Cali, Colombia\\
hbenitez@javerianacali.edu.co}
\and
\IEEEauthorblockN{Ivan Mauricio Cabezas}
\IEEEauthorblockA{\textit{School of Engineering} \\
\textit{Universidad de San Buenaventura}\\
Cali, Colombia\\
imcabezas@usbcali.edu.co
}
}

\maketitle

\begin{abstract}

Object tracking in realistic scenarios is a difficult problem
affected by various image factors such as occlusion, clutter, confusion, object shape, unstable speed, and zooming. While these conditions do affect tracking performance, there is no clear distinction between the scene dependent challenges like occlusion, clutter, etc., and the challenges imposed by traditional notions of impairments from capture, compression, processing, and transmission. This paper is concerned with the latter interpretation of quality as it affects video tracking performance. In this work we aim to evaluate two state-of-the-art trackers (STRUCK and TLD) systematically and experimentally in surveillance videos affected by in-capture distortions such as under-exposure and defocus. We evaluate these trackers with the area under curve (AUC) values of success plots and precision curves. In spite of the fact that STRUCK and TLD have ranked high in video tracking surveys. This study concludes that in-capture distortions severely affect the performance of these trackers. For this reason, the design and construction of a robust tracker with respect to these distortions remains an open question that can be answered by creating algorithms that makes use of perceptual features to compensate the degradations provided by these distortions.

\end{abstract}


\section{Introduction}

A significant number of video quality databases have been designed
in the last years \cite{Simone2010}, \cite{Moorthy2012}, \cite{Group}, \cite{Nuutinen2016}, \cite{Ghadiyaram2017a}, \cite{Ghadiyaram2014}. All of these databases have been developed by first obtaining a small set of high-quality videos, then systematically distorting them in a controlled manner. Furthermore, most of the existing tracking and VQA (Video Quality Assessment) datasets do not contain simultaneously in-capture and post-capture distortions or only have a single distortion type \cite{SeshadrinathanSoundararajanBovikEtAl2010}, \cite{caviar}. Moreover, they do not model in-capture authentic distortions \cite{SchuldtLaptevCaputo2004}. 

In \cite{Deepti_2017}, Deepti et al. present a recently constructed video database that contain in-captured distortions. In this work the designed database comprises a total of 208 videos that were captured using eight different smart-phones. The videos in this database contain six common in-capture distortions (artifacts, color, exposure, focus, sharpness, and stabilization).
The purpose of this study was to study how real-world in-capture distortions challenge both human viewers as well as automatic perceptual quality prediction models.

To the extent of our knowledge, very little work has been done on the construction  databases affected by in-capture distortions for video surveillance applications. For instance, Tsifouti et al. \cite{Tsifouti2012} generated degraded datasets that allow to test how video compression and frame rate reduction affects the performance of analytics systems. They concluded that the performance of the systems depends on the specific implementation of the software used for the compression, on the target bit rate and on the frame rate. Furthermore, they reported that the compression methods increased the false positives. They proposed as future work the analysis of properties such as low contrast ratio and low brightness/dark events that affect the performance of video analytics systems. Gao, Zhang et al. \cite{Zhang2014}, \cite{Zhang2014a}, \cite{Gao2014} generated the PKU-SVD-A\footnote{http://mlg.idm.pku.edu.cn/resources/pku-svd-a.html} dataset and conducted several experiments on these videos (1080p uncompressed). They resized to distinct resolutions and compressed with different quantization parameters for evaluating the effects of video compression on typical analysis tasks.

This paper describes the creation of a distorted video surveillance dataset affected by in-capture distortions and acquired with four different surveillance cameras. Hence, to analyze the impact of video distortions on state-of-the-art video trackers, it is necessary to design and develop video databases that contain scenes of interest for video tracking applications. The intended solution strategy to solve this issue is to create video sets with varied content of indoor and outdoor scenes of interest to test tracking algorithms. This dataset aims to be shared with the scientific community, through the creation of an open-access repository. This paper is organized as follows. 
Section II presents the characteristics of the state-of-the-art video trackers used to analyze the distorted surveillance videos. Part III describes the design of the experimental  setup for the acquisition of distorted videos from different four video cameras commonly used in surveillance applications and the specifications of additional datasets deployed to obtain tracking results.   

\section{Trackers TLD and STRUCK }

\subsection{TLD tracker}

TLD tracker  uses labeled and unlabeled samples for discriminative classifier learning. The method is applied to tracking by combining the results of a detector and an optical flow tracker. Given the target bounding box in the first frame, the detector learns an appearance model from binary patterns differentiated from patterns obtained from a distant background. The optical flow tracker applies a Lucas Kanade Tracker (KLT) \cite{Baker2004} to the target region and proposes a target window in the current frame. The normalized cross correlation is computed for the candidate windows. The system selects the candidate window which has the highest similarity to the object model as the new object. Once the target is localized, positive samples are selected in and around the target and negative samples are selected at further a distance to update the detector target model. If neither of the two trackers outputs a window, TLD declares loss of target. In this way TLD can effectively handle short-term occlusion. The code of this tracker was obtained from \cite{TLD}. 

\subsection{STRUCK tracker}

STRUCK stands for Structured output tracking with kernels. This algorithms carries out a discriminative tracking with constraints. 
The structured supervised classifier \cite{STRUCK} circumvents the acquisition of positively and negatively labeled data altogether, as it integrates the labeling procedure into the learner in a common framework. Given the target bounding box, it considers different windows by translation in the frame. Using these windows as input, the structured SVM (S-SVM) accepts training data of the form {appearance of the windows, translation}. The window’s appearance is described by Haar features arranged on a 4x4 grid and 2 scales, raw pixel intensities on a 16x16 rescaled image of the target and intensity histograms. In a new frame, candidate windows are sampled uniformly around the previous position. The classifier computes the corresponding discriminant highest score selected as the new target location. Subsequently, the new data for updating the S-SVM are derived from the new target location. While updating the S-SVM learner enforces the constraint that the current location still stays at the maximum. Locations which violate the constraint will become support vectors. In practice, as maximization during updating is expensive, the learner uses a coarse sampling strategy.

\section{Surveillance Video Databases}
 
Surveillance cameras acquired the videos with a ~10 fps  frame rate. The reason is that in commercial application of video surveillance the average value of the frame rate is ~10 fps \footnote{\url{https://ipvm.com/reports/frame-rate-surveillance-guide}}. This frame rate selection is motivated by the minimization of storage costs.
 
\subsection{Distorted Surveillance Video Database (DSVD)}

\begin{figure}[htbp]
	\centering
	 \includegraphics[width=.3\linewidth]{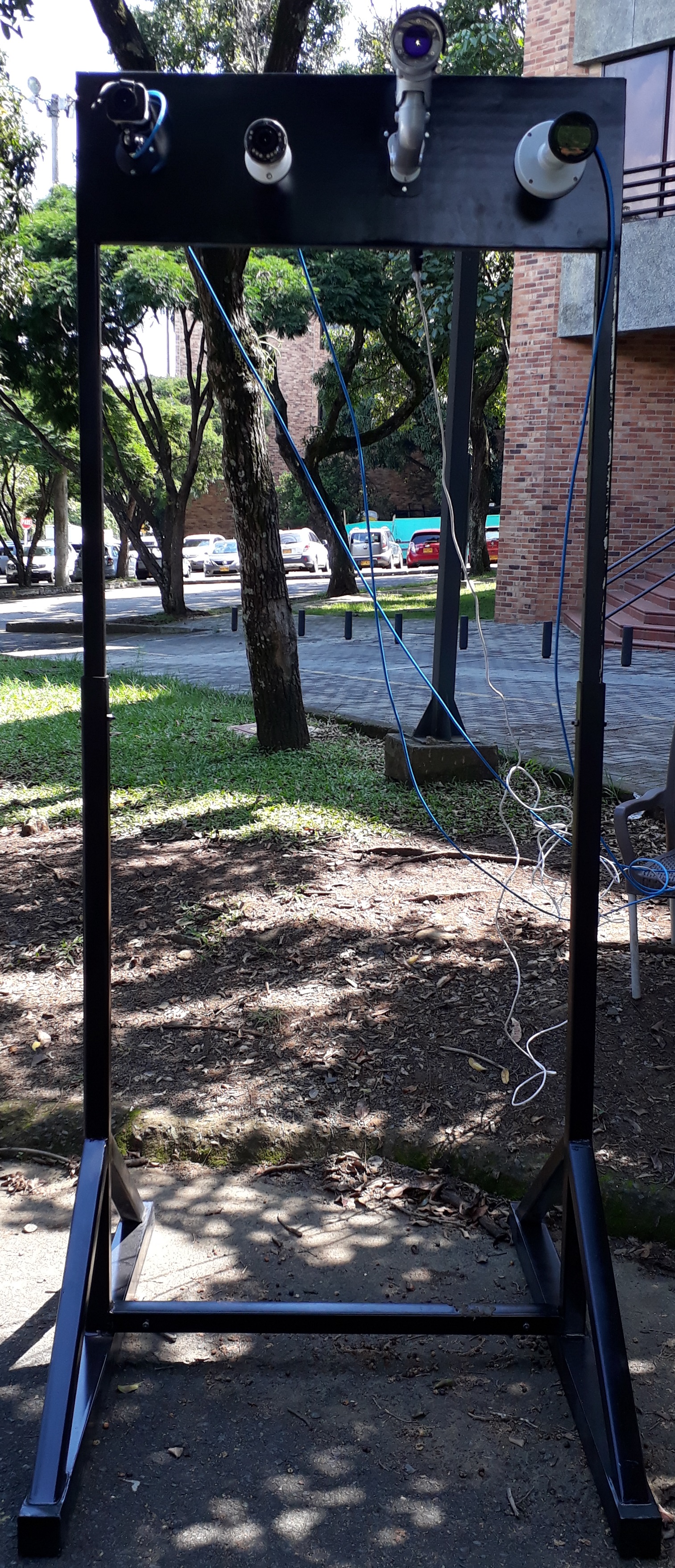}
		\caption{Structure used to support the cameras }
		\label{fig:structure_physical}
\end{figure}

For temporal synchronization, the recorded videos  have an equal rate I/P frames: 10 fps. Fig.~\ref{fig:structure_physical} shows the experimental set-up that support the surveillance cameras. The video cameras are synchronized by using a VMS and they are aligned in such a way that the fields of view overlaps as much as possible to acquire similar spatial-temporal information of the scenes. 

To create the dataset, a number of video clips are recorded. We deploy the structure shown in Fig.~\ref{fig:structure_physical}  as the physical support to sustain the surveillance cameras. The surveillance camera shown in Fig.~\ref{fig:structure_physical} are (from left to right): 

\begin{enumerate}
\item VIVOTEK IP8165 (C1)
\item VIVOTEK IB8367 (C2)
\item VIVOTEK IB8381 (C3)
\item AXIS \qquad P1405     \  (C4)
\end{enumerate}

\begin{figure}[htbp!]
\centering
\begin{subfigure}{.24\textwidth}
  \centering
  \includegraphics[width=1\linewidth]{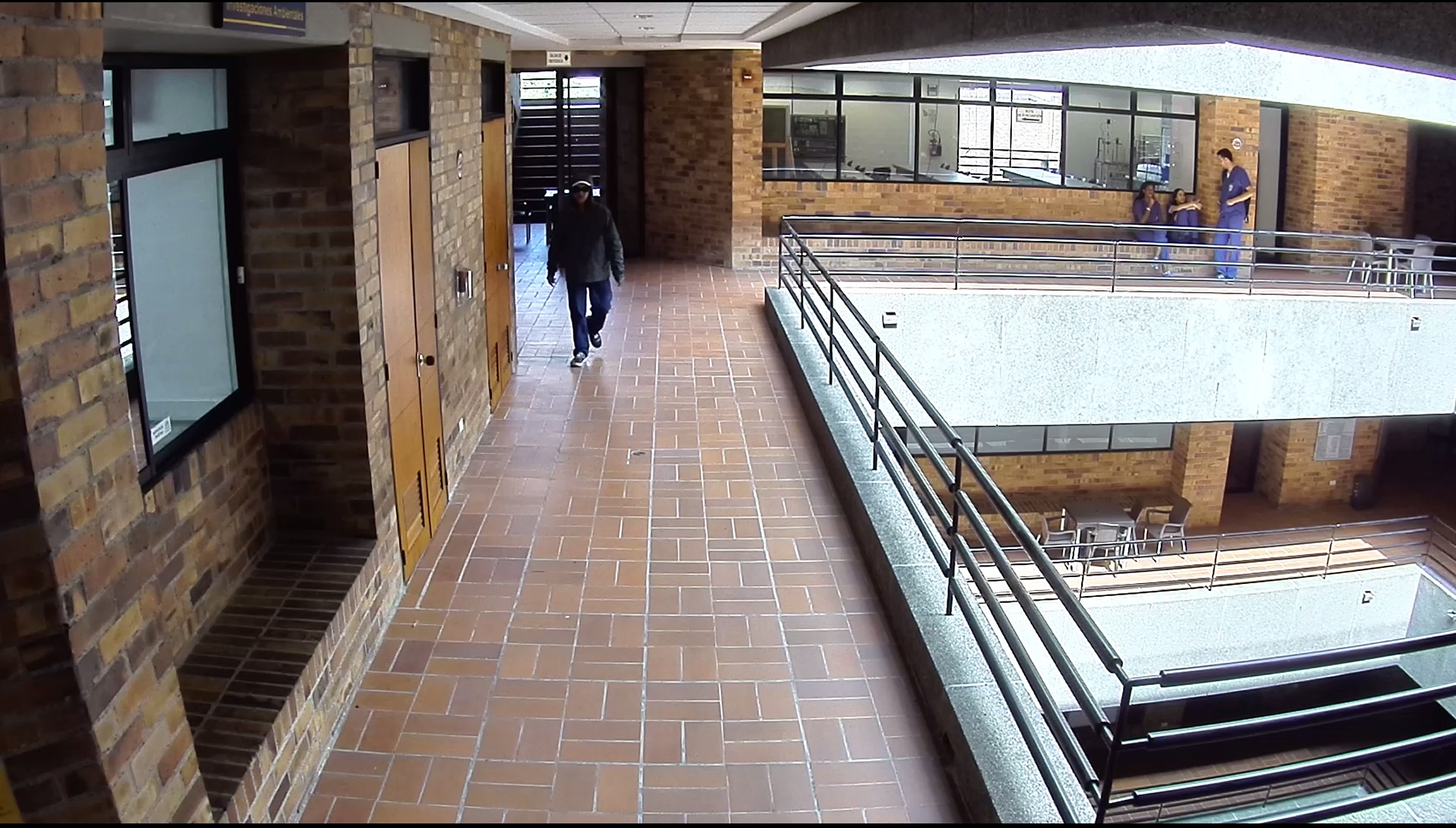}
  \caption{Original Image, nor H.264 compression or cropped.}
  \label{fig:sub1_ImageCompression}
\end{subfigure}%
\begin{subfigure}{.24\textwidth}
  \centering
  \includegraphics[width=1\linewidth]{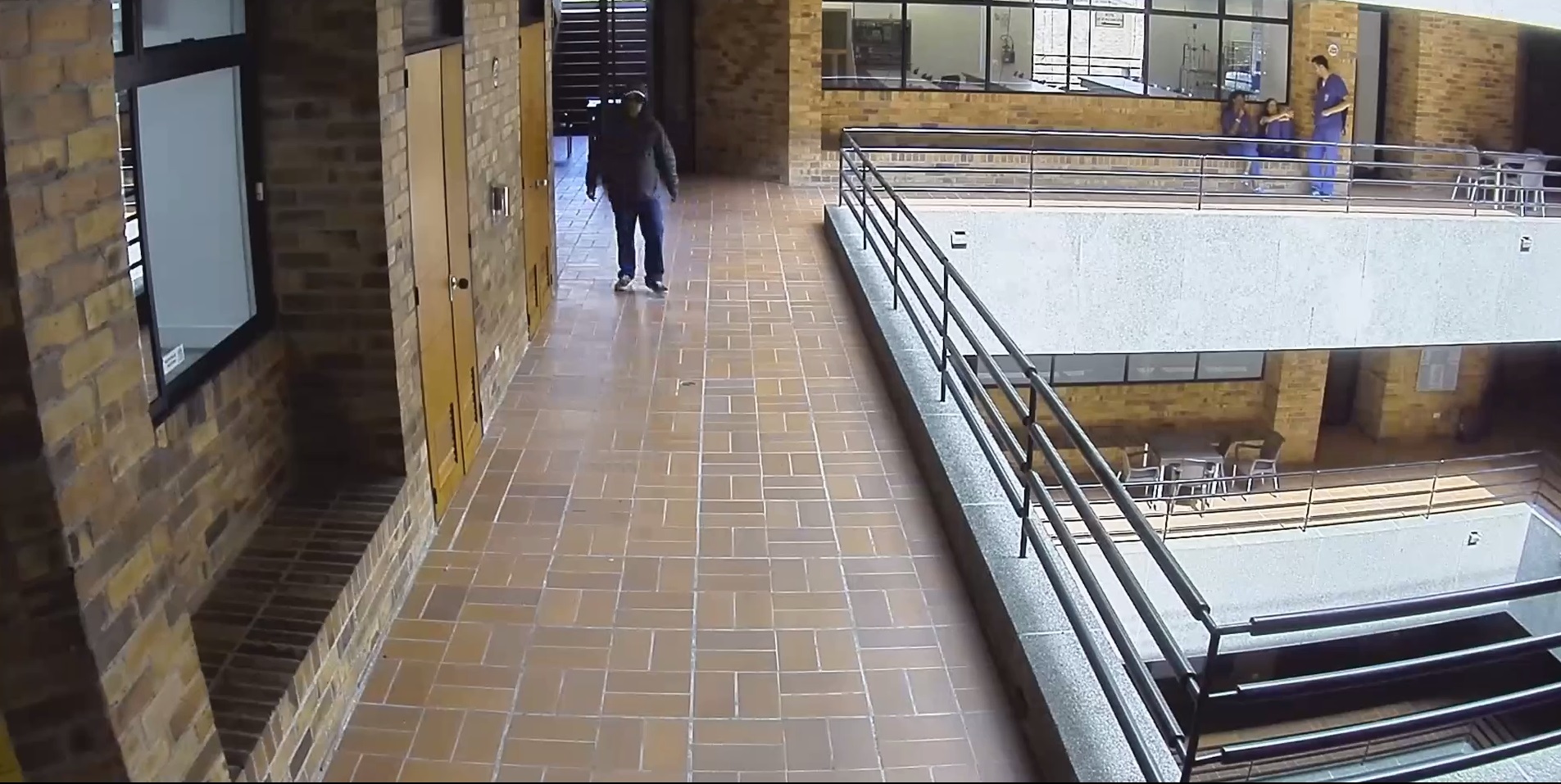}
  \caption{Image with compression 50 (Medium Quality)}
  \label{fig:sub2_ImageCompression}
\end{subfigure}
\begin{subfigure}{.24\textwidth}
	\centering
	\includegraphics[width=1\linewidth]{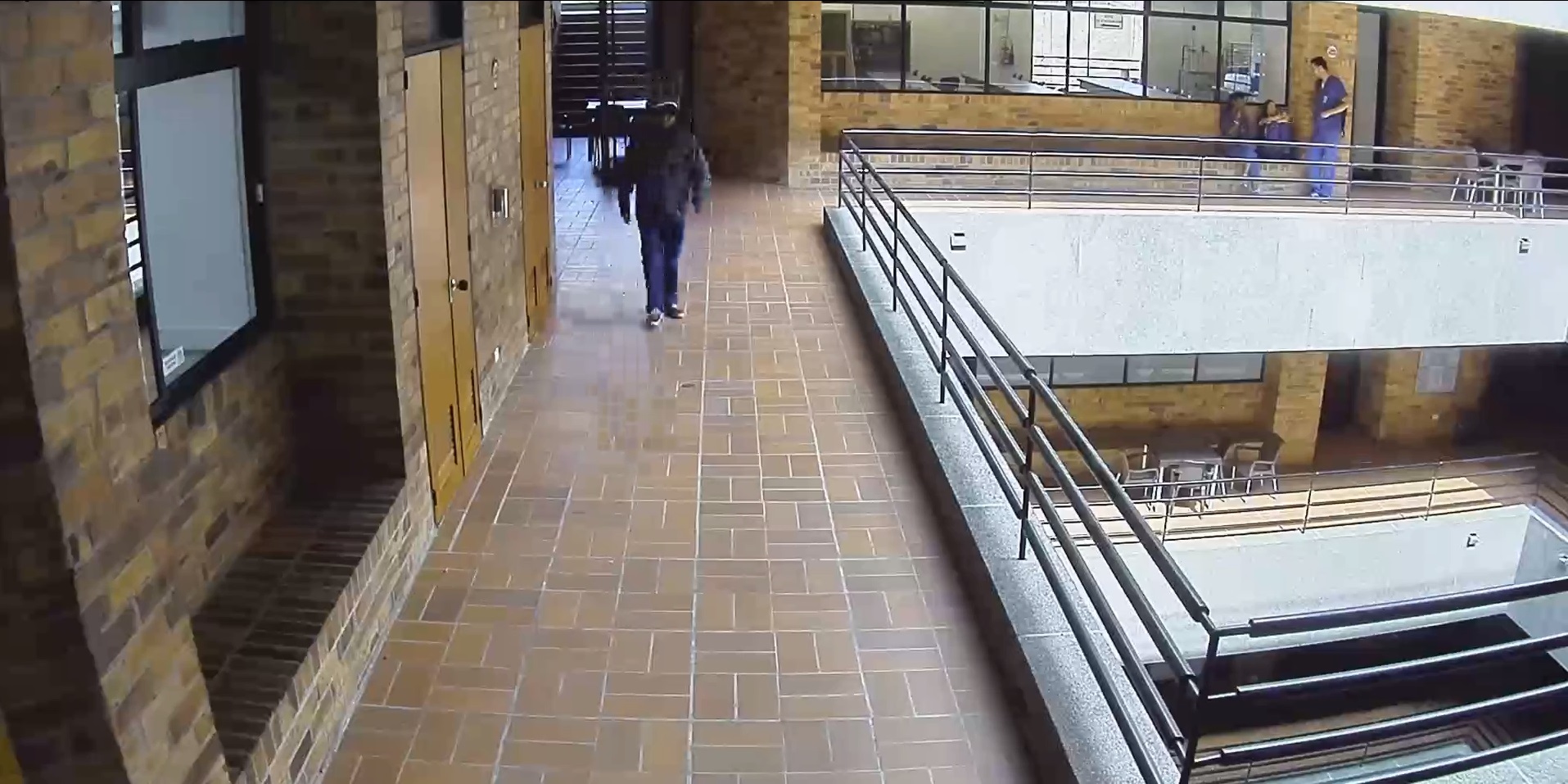}
	\caption{Image with compression factor =75 (low quality)}
	\label{fig:sub3_ImageCompression}
\end{subfigure}
\begin{subfigure}{.24\textwidth}
	\centering
	\includegraphics[width=1\linewidth]{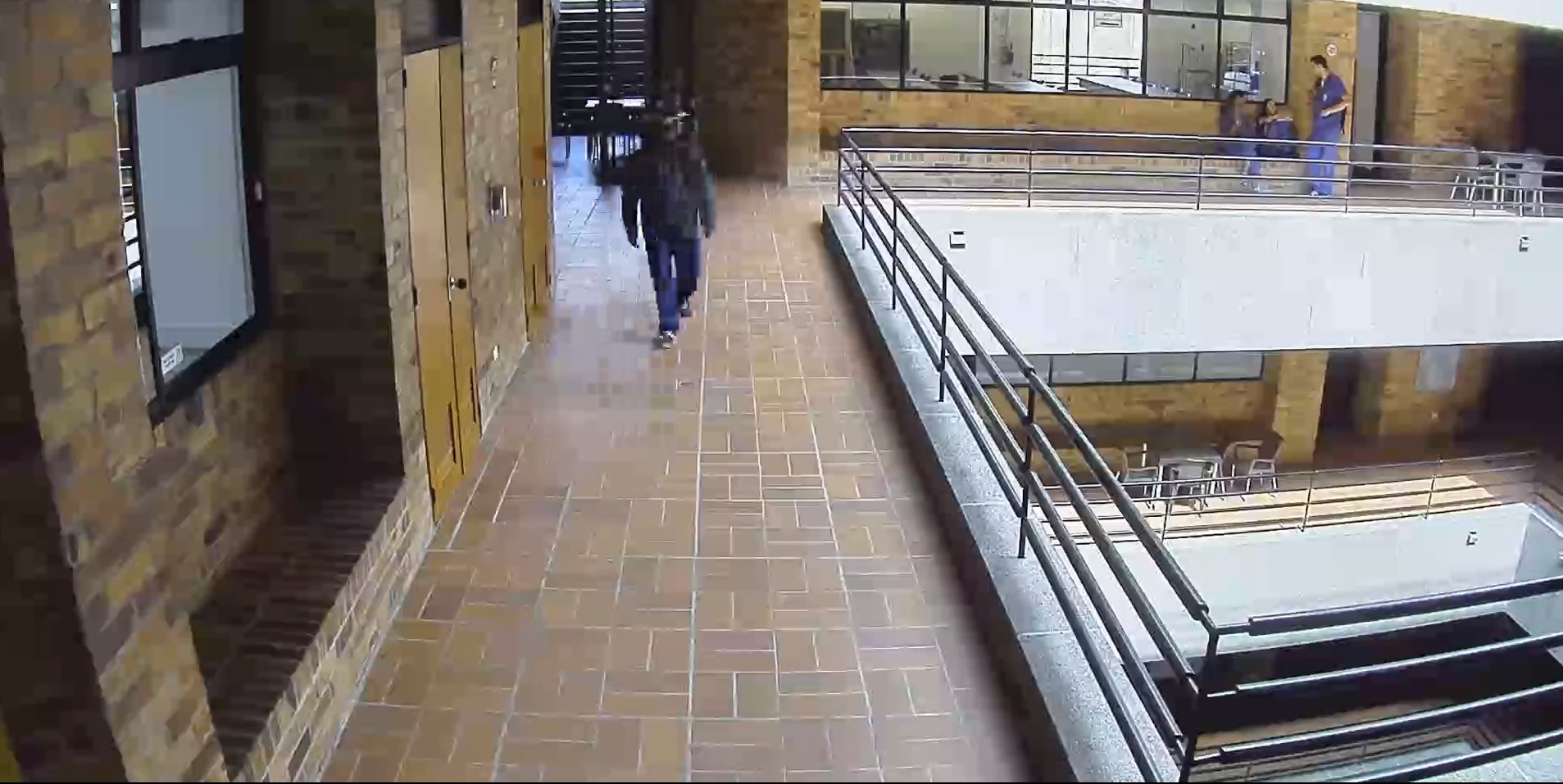}
	\caption{Image with compression factor =100 (lowest quality)}
	\label{fig:sub4_ImageComp016Exp_IndWL_FQ_C4_STRUCK_Resultsression}
\end{subfigure}
\caption{Images with different levels of distortion for compression H.264 (pos-capture)}
\label{fig:Images_Distortion_Compression}
\end{figure}

The VMS (Video Management Software, Digifort) compresses the video in the H.264 format with a time label in the upper right part of the frame. Since this timestamp is unnecessary for the video tracking, we decided to remove it in such a way that in the  compressed videos the height of the frame is reduced in 124 rows. The video sequences have been degraded by using a H.264/AVC compression scheme at three different bitrates, resulting in 3 mirrored video sequences, that differ only in the level of compression \cite{Ninassi2009}. The three different bitrates were chosen in order to generate degradations all over the distortion scale (from imperceptible to very annoying, as shown in Fig.~\ref{fig:Images_Distortion_Compression}.

\subsection{Tracking, Learning and Detection (TLD) dataset}

In this report we analyze three videos from the publicly available TLD database (Fig.~\ref{fig:TLD_motocross} - Fig.~\ref{fig:TLD_David}). These videos are motocross, jumping, and David. These videos present challenges for the trackers such as uneven illumination and changes of object velocity.

\subsection{Database of videos of real crime scenes}

\begin{figure}[htbp!]
\centering
\begin{subfigure}{.24\textwidth}
  \centering
  \includegraphics[width=1\linewidth]{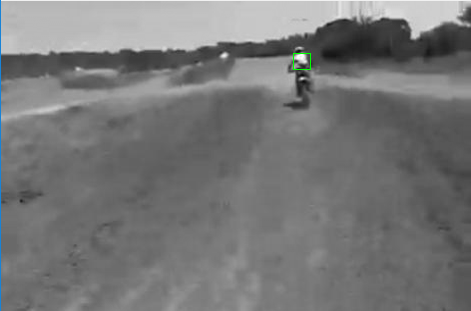}
  \caption{Image from TLD dataset motocross}
  \label{fig:TLD_motocross}
\end{subfigure}%
\begin{subfigure}{.24\textwidth}
  \centering
  \includegraphics[width=1\linewidth]{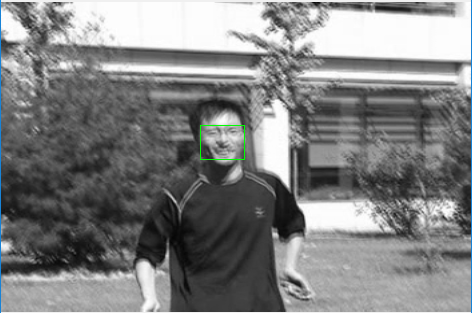}
  \caption{Image from TLD dataset jumping}
  \label{fig:TLD_Jumping}
\end{subfigure}
\begin{subfigure}{.24\textwidth}
	\centering
	\includegraphics[width=1\linewidth]{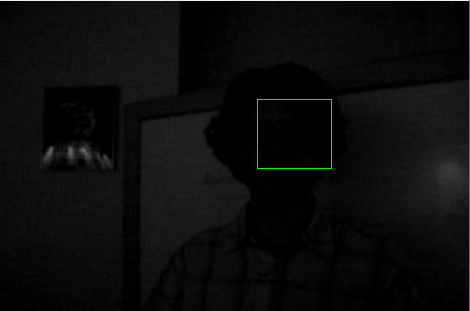}
	\caption{Image from TLD dataset David (Illumination changes).}
	\label{fig:TLD_David}
\end{subfigure}
\begin{subfigure}{.24\textwidth}
	\centering
	\includegraphics[width=1\linewidth]{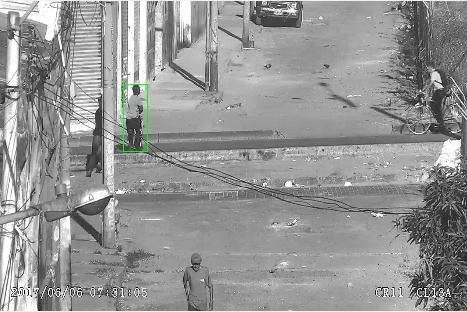}
	\caption{Image from police video surveillance}
	\label{fig:PoliceVideo}
\end{subfigure}
\caption{Images from different datasets used in the rotation.}
\label{fig:Dataset_Images}
\end{figure}

The local Police Department in Santiago de Cali has a video surveillance camera network composed of approximately 1500 units located in different region of the city. In this report the video trackers were tested with one video of a bike robbery in Cali streets (Fig.~\ref{fig:PoliceVideo}). This video is taken from a set of videos that we name 
local police surveillance videos (LPSV) and include challenging conditions for the trackers such as occlusion, scale changes, camera motion, and object getting out of scene.

\section{Results}

The measures that evaluates the  video trackers performance are
success rate and precision plots. Success rate is obtained by finding the percentage of pixels overlap between the tracker output and the ground truth bounding Box. The overlap scores determines whether an algorithm successfully tracks a target object in one frame, by testing whether this overlap is larger than a given threshold. The average success rate with a threshold fixed to $25\%$, is used here for the performance evaluation  \cite{Babenko2011}. Precision plot is the center location error, which computes the average euclidean distance between the center locations of the tracked targets and the ground truth.

\begin{figure}[htbp!]
\centering
\begin{subfigure}{.24\textwidth}
  \centering
  \includegraphics[width=1\linewidth]{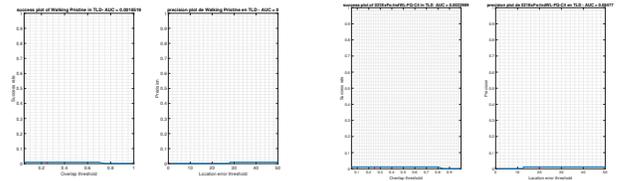}
	\caption{TLD results for walking person, pristine video.}
  \label{fig:TLD_Pristine_Walking}
\end{subfigure}%
\begin{subfigure}{.24\textwidth}
  \centering
  \includegraphics[width=1\linewidth]{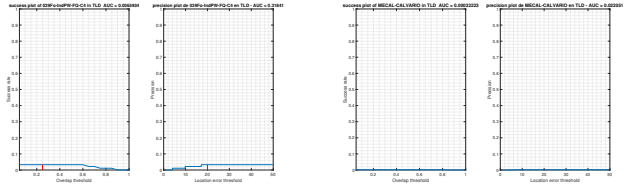}
	\caption{TLD for walking person, with exposure and focus distortion.}
  \label{fig:TLD_Exposure_Focus}
\end{subfigure}
\begin{subfigure}{.24\textwidth}
	\centering
	\includegraphics[width=1\linewidth]{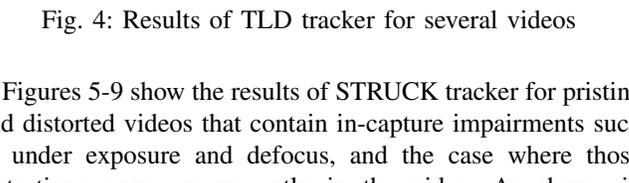}
	\caption{TLD results for walking person, with focus distortion.}
	\label{fig:TLD_Focus}
\end{subfigure}
\begin{subfigure}{.24\textwidth}
	\centering
	\includegraphics[width=1\linewidth]{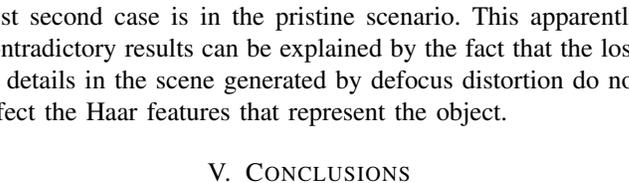}
	\caption{TLD results for police real Video.}
	\label{fig:TLD_PoliceVideo}
\end{subfigure}
\caption{Results of TLD tracker for several videos}
\label{fig:TLD_Results}
\end{figure}

According to the success plots AUC curves in Figure \ref{fig:TLD_Results} the TLD tracker has a low performance in videos such as David and Motocross that present low illumination, high velocity, and scale changes. For the databases LPSV and DSVD,the TLD performance expressed by the area under curve (AUC) for success curves and precision plots is less than 0.05. This result reflects a poor performance of TLD tracker in these databases affected by distortions such as under exposure and defocus. 

\begin{figure}[htbp!]
	\centering
	\includegraphics[width=0.49\textwidth]{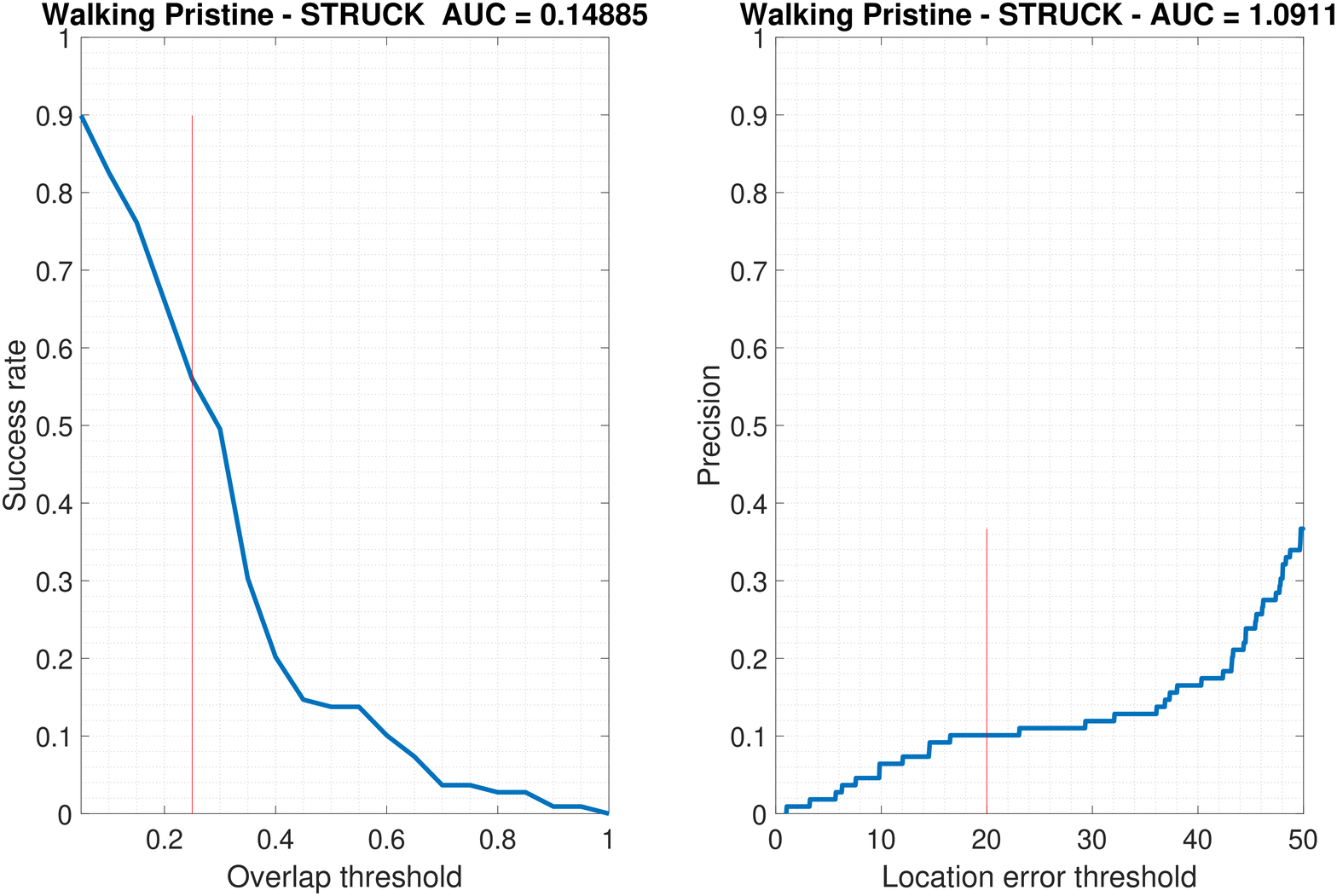}
	\caption{STRUCK results for walking person pristine video}
	\label{Fig:WLPristine_Struck}
\end{figure}

\begin{figure}[htbp!]
	\centering
	\includegraphics[width=0.49\textwidth]{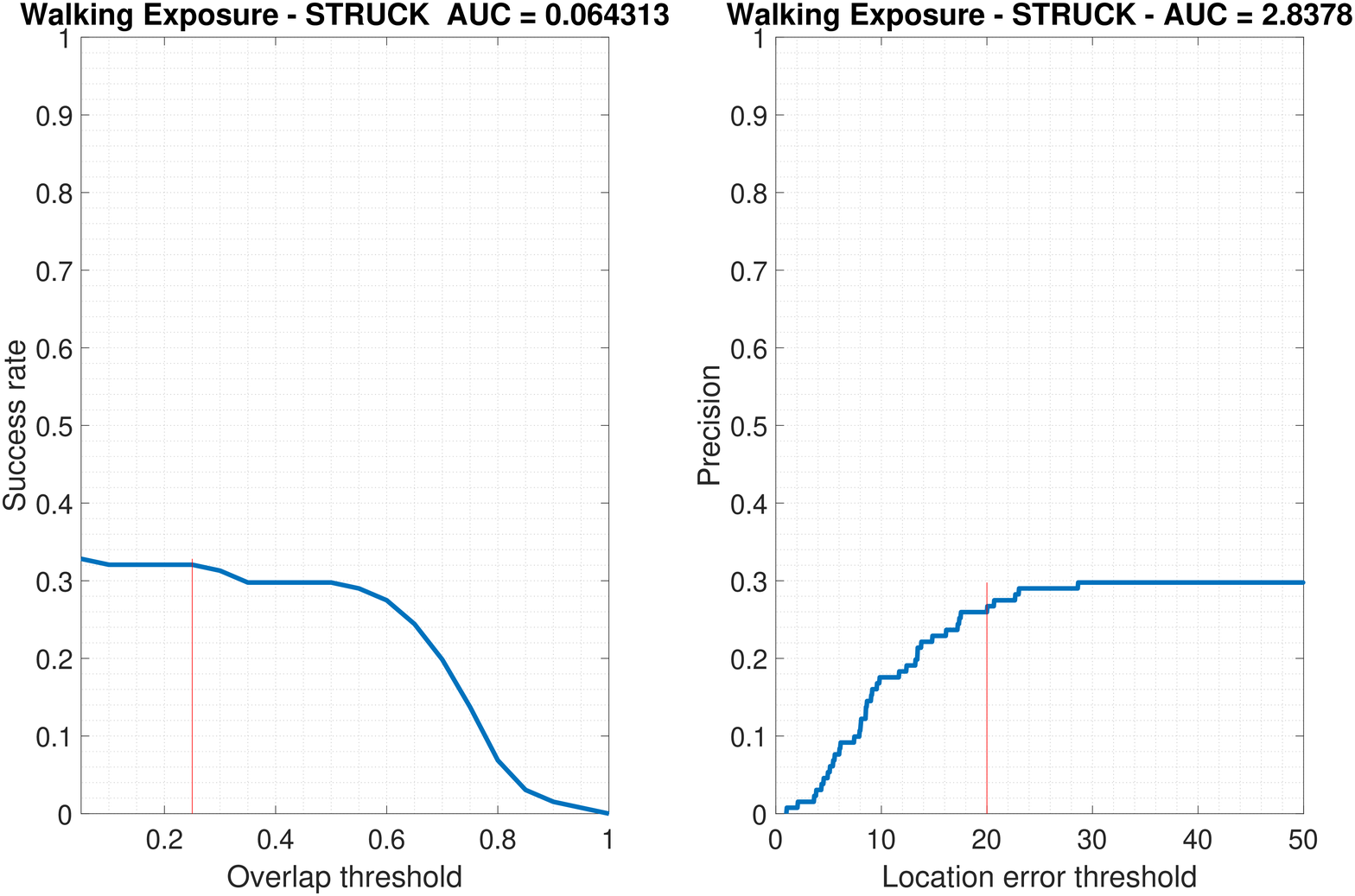}
	\caption{STRUCK results for walking person in exposure distorted video. Exposure = $\frac{1}{480}$}
	\label{Fig:WL_Exposure_StruckResults.PNG}
\end{figure}

\begin{figure}[htbp!]
	\centering
	\includegraphics[width=0.49\textwidth]{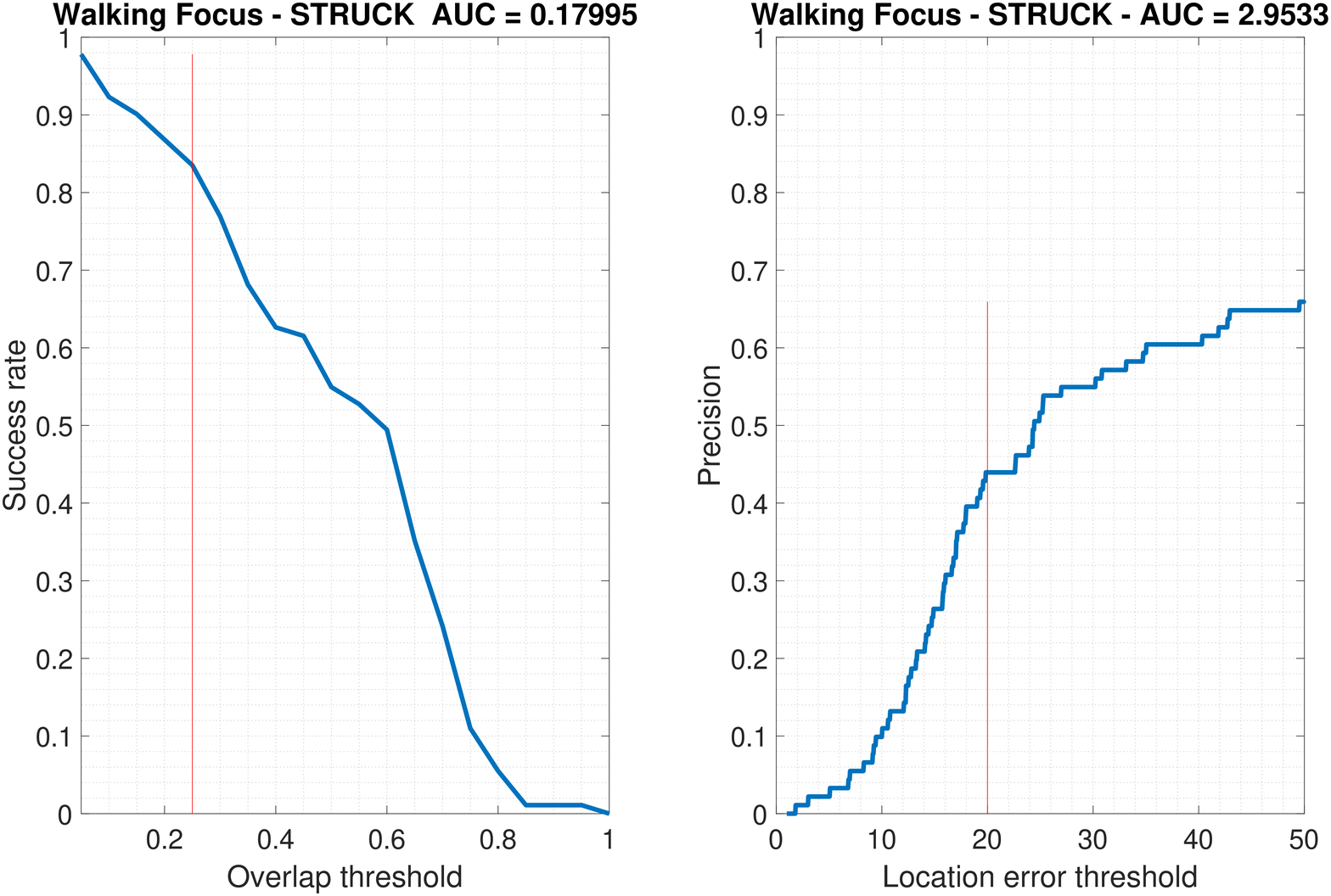}
	\caption{STRUCK results for walking person in focus distorted video. Focus deviation = 10}
	\label{Fig:WLFo_StruckResults.PNG}
\end{figure}

\begin{figure}[htbp!]
	\centering
	\includegraphics[width=0.49\textwidth]{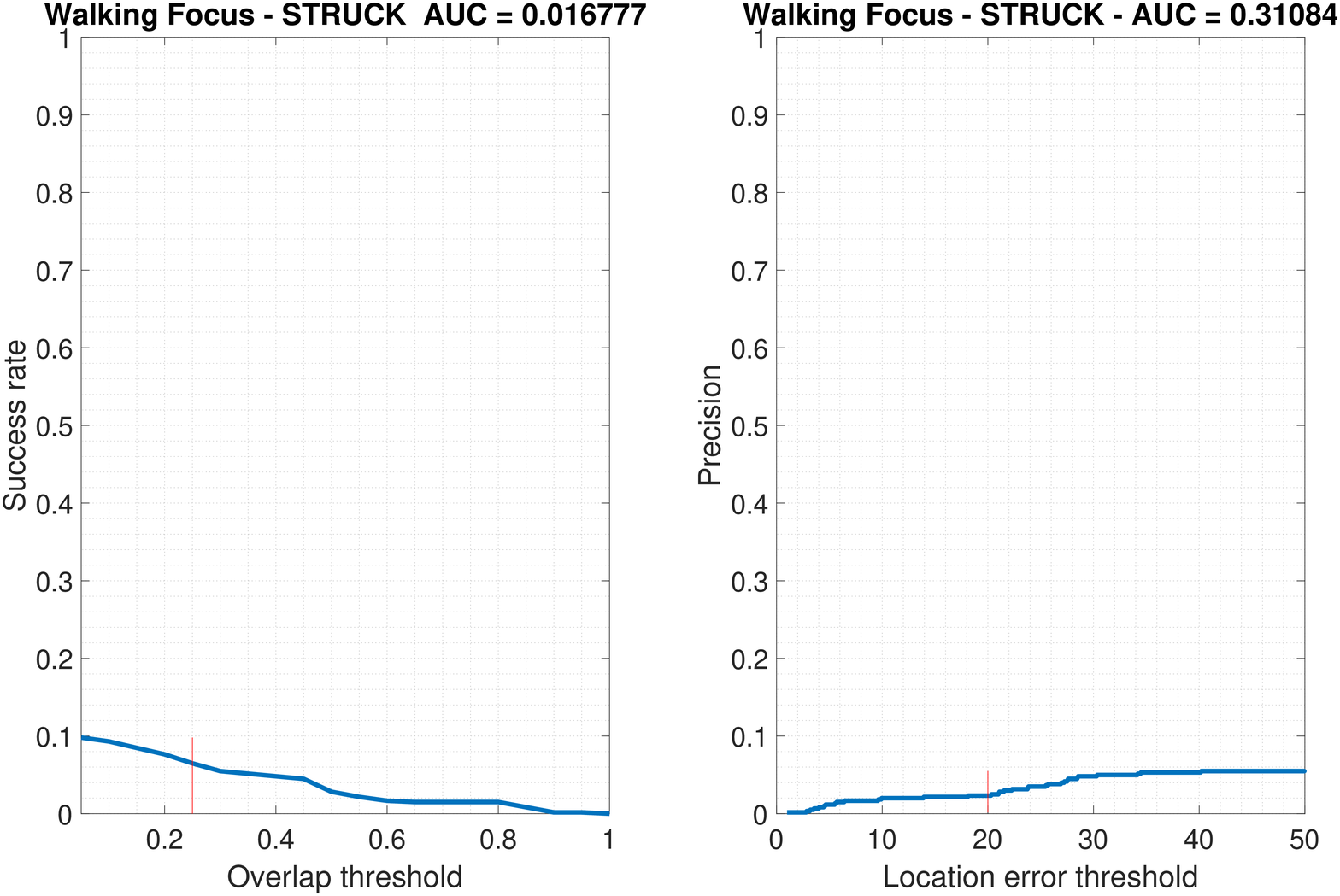}
	\caption{STRUCK results for police video surveillance }
	\label{Fig:MECAL_Calvario_Struck.PNG}
\end{figure}

Figures \ref{Fig:WLPristine_Struck}-\ref{Fig:MECAL_Calvario_Struck.PNG} show the results of STRUCK tracker for pristine and distorted videos that contain in-capture impairments such as under exposure and defocus, and the case where
those distortions occur concurrently in the video. As shown in figures \ref{Fig:WLPristine_Struck} to \ref{Fig:MECAL_Calvario_Struck.PNG}, STRUCK obtained the best tracking results when the defocus distortion reached the most severe level. The best second case is in the pristine scenario. This apparently contradictory results can be explained by the fact that the loss of details in the scene generated by defocus distortion do not affect the Haar features that represent the object. 

\begin{figure}[htbp!]
	\centering
	\includegraphics[width=0.49\textwidth]{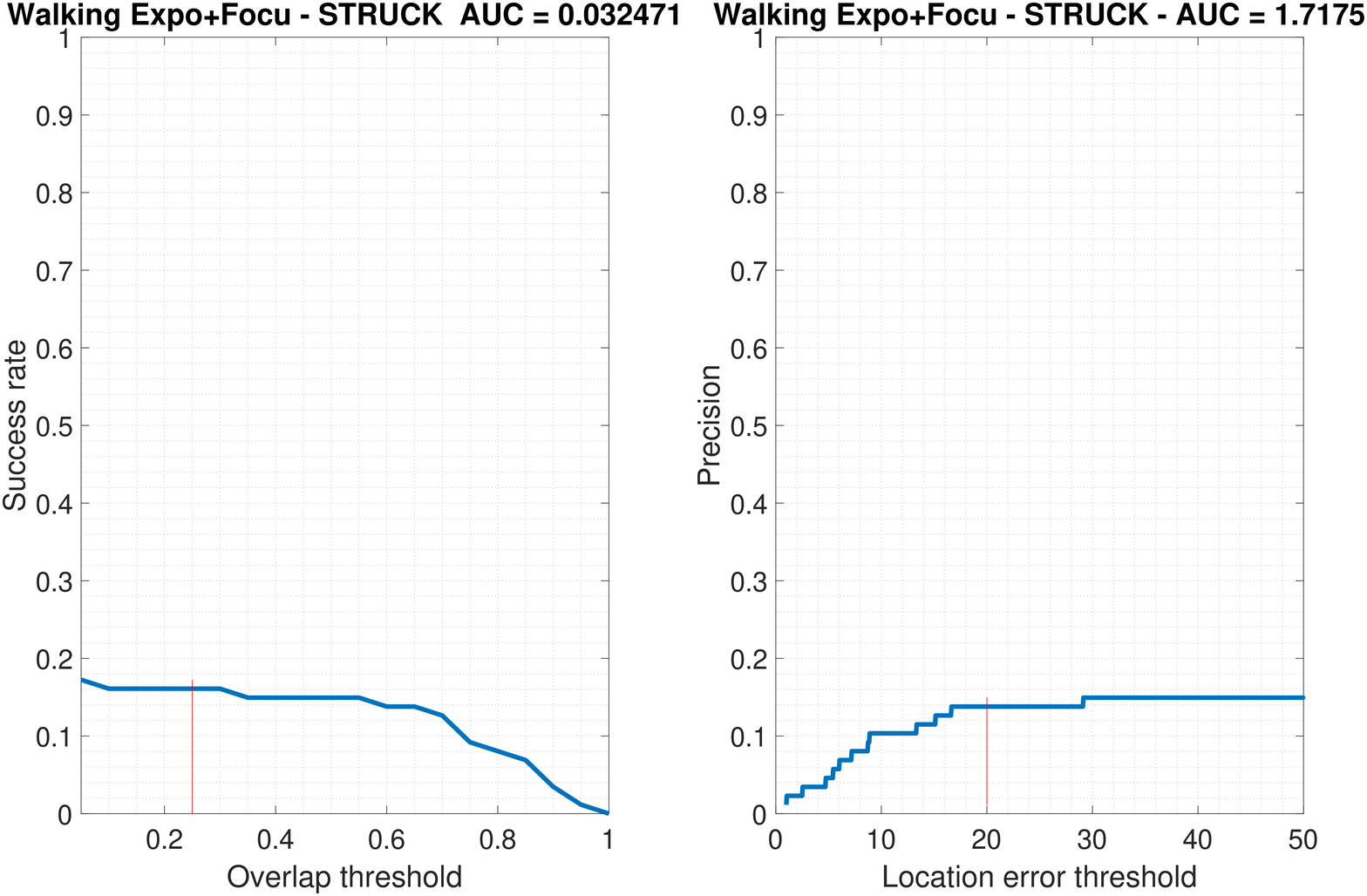}
	\caption{STRUCK results for walking person in exposure and focus distorted video. Exposure = $\frac{1}{120}$, Focus deviation = 10}
\end{figure}

\section{Conclusions}

We propose a procedure to change in a controlled way the exposure time and focus of four different commonly used surveillance cameras. We built a controlled experimental set-up composed of four surveillance cameras, a switch, and a VMS  to acquire videos affected by in-capture distortions. This is the first attempt to create a Distorted Video Surveillance Database DVSD (publicly available in \url{https://goo.gl/1o3FbW}) that contain videos affected by in-capture distortions produced by exposure time and defocus variations. This database is a solid starting point to study the influence of distortions on video tracker performance.  

Furthermore, we designed a set of experiments at different scenarios and locations to acquire distorted videos that contain scenes of interest with activities such as  people walking alone, meeting with others, fighting and passing out, leaving a package in a public place,  prowling, and robbery. We carried out an analysis of the state-of-the-art trackers STRUCK and TLD in new challenging scenarios that include distorted videos with in-capture distortions and real world surveillance scenes. 

In spite of the fact that STRUCK and TLD have ranked high in the survey studies carried out in \cite{Smeulders_Survey} and \cite{Wu2015}, this study concludes that in-capture distortions severely affect the performance of these trackers. For this reason, the design and construction of a robust tracker with respect to these distortions remains an open question that can be answered by creating algorithms that makes use of perceptual features to compensate the degradations provided by these distortions. 

\clearpage

\bibliographystyle{plain}
\bibliography{biblist}

\end{document}